\pdfoutput=1

\documentclass[11pt]{article}

\usepackage{acl}

\usepackage{times}
\usepackage{latexsym}

\usepackage[T1]{fontenc}

\usepackage[utf8]{inputenc}

\usepackage{microtype}

\usepackage{graphicx}

\usepackage{amsmath}

\title{A Human Subject Study of Named Entity Recognition (NER) in Conversational Music Recommendation Queries}

\author{Elena V. Epure \and Romain Hennequin \\
    Deezer Research, Paris, France \\ \texttt{research@deezer.com}}

\begin{document}

\maketitle
\begin{abstract}
We conducted a human subject study of named entity recognition on a noisy corpus of conversational music recommendation queries, with many irregular and novel named entities.
We evaluated the human NER linguistic behaviour in these challenging conditions and compared it with the most common NER systems nowadays, fine-tuned transformers.
Our goal was to learn about the task to guide the design of better evaluation methods and NER algorithms.
The results showed that NER in our context was quite hard for both human and algorithms under a strict evaluation schema;
humans had higher precision, while the model higher recall because of entity exposure especially during pre-training; and entity types had different error patterns (e.g. frequent typing errors for artists).
The released corpus goes beyond predefined frames of interaction and can support future work in conversational music recommendation.
\end{abstract}

\section{Introduction}
\label{sec:introduction}
Music recommendation systems (RSs), fundamental to streaming services nowadays, learn from user listening history or music content which artists or tracks to suggest next \citep{schedl2018current}.
Most of these algorithms provide personalized music content to the users when logging in the streaming apps or websites, or when triggered with pre-defined utterances via voice assistants \citep{Ammari2019,Bontempelli2022}. 
More recent \textit{conversational} RSs aim to help users to express their recommendation needs by supporting interactions via \textit{queries} in natural language \citep{Jannach2021survey}.
However, despite existing in the scientific literature, such conversational RSs are not widely deployed because of multiple issues, one being NER.

The processing of recommendation queries entails the \textit{extraction of named entity mentions} \citep{moon-etal-2019-opendialkg,Rongali2020}. 
This sub-task faces multiple challenges, even when queries are framed as pre-defined utterances.
The transcriptions of the voice queries results in lower-case noisy text, often with misspellings \citep{muralidharan-etal-2021-noise}.
The lack of capitalisation in entities and misspelled words are often present in text-based queries too \citep{Cheng2021}.
Music entities, or those coming from the creative content domains, are highly irregular:
they do not follow inherent patterns as it is the case with people's names, and there is little to no separation between the vocabularies of entity and context words, especially for creative works \citep{derczynski-etal-2017-results} (e.g. common words like "I" or "love" in track titles).
Also, new music entities appear all the time.
Major music streaming services ingest one new track almost every second \citep{Ingham2021}.

Previous works have already shown that NER systems struggle with the aforementioned challenges \citep{Augenstein2017,lin-etal-2020-rigorous,epure2022probing}.
Thus, multiple approaches have been proposed to address them, either focused 1) on collecting more and relevant data for training / fine-tuning standard NER sequential models \citep{lison-etal-2020-named}; or 
2) on model's design choices that favour generalisation \citep{guerini-etal-2018-toward,lin-etal-2020-triggerner}.
Most solutions focused on the latter objective have been motivated by the \textit{human NER linguistic behaviour}, e.g. make the model rely more on context cues than on named entity mentions or learn from a few examples only, as humans do.
However, apart from some scarce, partially related works \citep{derczynski-etal-2016-broad,ding-etal-2021-nerd}, there is no systematic investigation of how humans actually perform NER on noisy text with many new and irregular named entities.
Moreover, in the case of music recommendation, we are not aware of any existing dataset of queries in natural language, annotated with named entities.

Thus, our goal is to investigate the human NER linguistic behavior when confronted with these challenging conditions.
For that, we create \textit{MusicRecoNER}, a new corpus of noisy natural language queries for music recommendation in English that simulates human-music assistant interactions.
We then conduct a \textit{human subject research study} to establish a human baseline and learn from it.
Finally, we perform a detailed comparison of humans and the most popular NER systems nowadays, fine-tuned transformers, that covers multiple evaluation schemes (\textit{strict} named entity segmentation and typing, \textit{exact} segmentation only, or partial segmentation with strict named entity typing) and scenarios including entities previously seen or unseen by the model or humans.

The results showed that the task was challenging for humans.
Given an aggregated metric such as F1 score, human and algorithmic performances were on par.
However, the detailed evaluation revealed that humans struggled more with recall while the best model with precision.
The high recall obtained by the model was partially a result of entity exposure during pre-training or fine-tuning.
Also, music entities had different error patterns and, in some queries, had ambiguous context that made their segmentation and typing quite hard.

To sum up, our research contribution\footnote{Code and data are available at \url{https://github.com/deezer/music-ner-eacl2023}.} are:
\begin{enumerate}
    \item \textit{MusicRecoNER}, a corpus of noisy complex natural language queries for music recommendation collected from human-human conversations in English, but which simulates human-music assistant interactions, annotated with \textit{Artist} and \textit{WoA} (work of art) entities. This dataset is not limited to pre-defined utterances as it would be the case if collected from interactions with conversational or voice assistants. Thus, it contains entities in diverse context, being also a useful resource for future work on conversational music recommendation. 
    \item A \textit{human subject study design for NER} in noisy text with many new and irregular named entities. The proposed method is transferable to other creative content domains that face similar challenges to music such as books, movies, videos, but also to any other domain with scarce data, which wants to learn more about the NER task before building a system.
    
    \item An \textit{extensive music NER benchmark on noisy text} which compares the performance of human versus automatic baselines under multiple evaluation schemes, scenarios and by controlling for the \textit{novelty} of named entities.
\end{enumerate}

\section{Related Work}
\label{sec:relatedword}
Analysing human and algorithmic performance was done for multiple NLP tasks in the past.
\citet{nangia-bowman-2019-human} ran an annotation campaign on the GLUE benchmark with the goal to estimate the effort needed by existing models to catch up with the humans under limited-data regimes.  
\citet{kazantseva-szpakowicz-2012-topical} conducted a large-scale human study on topic shift identification in order to discover patterns of disagreements and consolidate the evaluation metrics.
\citet{ghaly-mandel-2017-analyzing} analysed the human behaviour for understanding ambiguous text-based or spoken sentences to guide the development of a machine learning system.
Multiple machine translation works challenged the human parity claim \citep{Toral2020} and proposed a secondary evaluation method to reveal detailed differences between humans or algorithms \citep{graham-etal-2020-assessing}.

Compared to these, we benchmark humans and models on a different task---named entity recognition, 
but we share similar goals---to estimate the human-algorithmic performance gap and to identify patterns that could support the design of better evaluation methods or automatic solutions.
Human annotation is frequent in NER especially when targeting a new domain such as archaeology \citep{brandsen-etal-2020-creating}, or a new language such as Indonesian \citep{khairunnisa-etal-2020-towards}.
However, we are not aware of any annotated corpus of conversational queries for recommendation in the music domain.
Some other related works propose corpora of noisy social media text containing new entities including irregular ones \citep{derczynski-etal-2016-broad,derczynski-etal-2017-results}, a noisy dataset of movie-related queries \citep{liu2014a}, a dataset of music artist biographies annotated for entity linking \citep{oramas-etal-2016-elmd}, or a corpus of tweets associated with a classical music radio channel \citep{Porcaro2019}.

Previous works have showed that transformers fine-tuned for NER are strong baselines, especially when training data is scarce \citep{akbik-etal-2019-pooled,fu2020rethinking}.
A more recent line of research employs these pre-trained models as few-short learners \citep{yang-katiyar-2020-simple,tanzer-etal-2022-memorisation}.
However, the results are still below those obtained with a fine-tuning approach.
In order to improve the bare-bone fine-tuned transformers, other works adopted distant supervision \citep{lison-etal-2020-named}, and the inclusion of gazetteers \citep{shang-etal-2018-learning} or contextual triggers \cite{lin-etal-2020-triggerner}.
Though these solutions are interesting and relevant to our problem and context, in the current research, we want to rely on the results of this study before making any design choices for an advanced NER system in the music domain.

When conducting human subject studies, the quality of annotations (inter-rater agreement or reliability) is often assessed with Kappa statistic or its variations \citep{mchugh2012interrater}.
Yet, for NER, or more generally for labelling phrases, this statistic is less applicable as the number of negative cases on which it relies is ill-defined \citep{hripcsak2005agreement}.
To address this issue, multiple imperfect solutions have been proposed such as to compute the Kappa statistic at the token level \cite{deleger2012building}---however, this does not reflect the task well as each token is not tagged individually; or to estimate the negative cases by enumerating all n-grams or noun phrases from a text---however, this lacks accuracy \citep{grouin-etal-2011-proposal}.
\citet{hripcsak2005agreement} show that when the number of negative cases gets very large, the Kappa statistic approaches the F1 score.
Thus, F1 is considered a better metric, which we also adopt to measure the performance of humans and compare the NER human and algorithmic baselines.

\section{Human Subject NER Study}
\label{sec:corpus}
\subsection{Data Collection}
\label{sec:datacollection}
For data collection we have chosen the \emph{music suggestions} subreddit\footnote{\url{www.reddit.com/r/musicsuggestions/}} as a relevant data source.
Reddit is a discussion website where members can submit questions, share content and interact with other members. 
It is organised in subreddits built around dedicated topics.
Each discussion starts with an initial post that has a title and description. 
From this post, threads of conversations develop.
We were interested only in posts triggered by a music information seeking or recommendation need.
We  crawled the full subreddit with 8615 initial posts.
This number corresponds to the posts in the beginning of 2020. 
We did not consider posts' comments. 

These humans-to-humans posts asking for music recommendations are particularly relevant to study as they go beyond pre-defined frames of interaction with a text or voice-based assistant.
Hence, they exhibit a realistic human use of language, which although more challenging, could help with the development of the next generation of music assistants. 
For NER, the existence of queries in natural language translates in a more diverse context surrounding named entities, thus in a higher query generalisation for music recommendation.
By manually checking this data, we noticed that many mentioned artists or music titles were not popular.
Thus, we expected most named entities to be new to the annotators, an aspect we wanted to control for, as mentioned in Section \ref{sec:introduction}.
  
\subsection{Data Cleaning and Pre-processing}
\label{sec:datapreprocessing}
As we aimed at creating a corpus of music recommendation queries simulating human-assistant interactions, we made multiple decisions to pre-process the collected posts.
We performed a manual cleaning of this data by removing those posts which directly shared music with the community; were aimed at promoting music or other music-related entities; contained explicit words; or contained only links to external music resources.

Then, we focused on \textit{titles} only as the post content was rather long, specific to asynchronous communication; 
as human-assistant interactions happen synchronously, the written or spoken queries are expected to be short, composed of a few short sentences at most \citep{Song2010}.
We removed all references to specific music-related services in order to obtain generic queries (e.g. we removed "Youtube" from the request "music similar to my Youtube playlist"). 
We also removed words which were explicit markers of human-human interaction in order to ensure compatibility with human-assistant interaction. For instance, we removed phrases such as "hello guys" or "could anybody".

\begin{table}
\begin{small}
    \centering
    \begin{tabular}{c|l}
        1 & looking for some playlists to listen to before going \\
        & to sleep i usually listen to beach house madlib etc \\ \hline
        2 &  ive just started listening to grateful dead and the \\
        &  ramones what else have i missed \\ \hline
       3 & looking for music similar to yamashita \\ \hline
       4 &   songs sounds like drive by lil peep\\ \hline
        5 &  new rappers\\
    \end{tabular}
    \caption{Examples of queries in \textit{MusicRecoNER}.}
    \label{tab:examples}
\end{small}
\end{table}

We performed the rest of the pre-processing steps to ensure that the queries contained, to some extent, the kind of noise that could be found in \textit{transcribed} voice queries too, such as those obtained when interacting with a voice assistant.
For this, we transformed the text in lowercase and removed punctuation marks and emoticons (with some exceptions when the symbol was part of the named entity's pronunciation such as "\&"). We kept content from parentheses when found at the end of a post title, otherwise we removed it. 
Although very common in automatic transcriptions, we did not introduce any artificial noise regarding the spelling of named entities.
Still some noise was present as Reddit authors sometimes made misspelling errors.
These steps were done automatically. 
We release both the original and pre-processed data.
All keywords used in the described steps are in Appendix \ref{app:filterkeywords}.
We show multiple query examples in Table \ref{tab:examples}.

\subsection{Annotation Guidelines and Procedure}
\label{sec:humanner}
We sampled multiple subsets of 600 queries each from the cleaned and pre-processed corpus.
This number was established by estimating the required time for the experiment to be maximum 2 hours per annotator, based on an initial trial on $751$ queries. 
The annotation guidelines were also tested in the trial experiment and refined after.
The subjects were informed that the goal was to identify names of artists (e.g. bands, singers, composers) and titles of works of art (e.g. albums, tracks, playlists, soundtracks) in unformatted music-related queries. 
We requested the annotators not to consult the Internet as we wanted them to rely on the query content only and on their own previous knowledge.

We then introduced the labels: \textit{Artist\_known}, \textit{Artist\_deduced}, \textit{WoA\_known}, \textit{WoA\_deduced}, and \textit{Artist\_or\_WoA\_deduced} with examples.
The last one was for ambiguous cases of named entity typing, but allowed the annotators to segment.
Segmentation is still very relevant when parsing natural language queries for music recommendation as the type could be eventually disambiguated with the help of a search engine, for instance.
The other labels corresponded to \textit{Artist} and \textit{WoA} types, completed by whether the annotator knew the entity from before or deduced it from query's content, as we wanted to keep track of \textit{entity's novelty}.

Then, we introduced challenging annotation cases with guidelines on how to proceed.
We instructed the annotators to include \textit{Artist} and \textit{WoA} named entities from other domains too such as movies or video games, but to ignore all the other entity types such as countries or music genres; 
to consider the innermost entities in case of nested entities; 
to ignore implicit entities such as "this singer";
to always include the "'s" from the possessive case as part of the named entity;
and to consider a named entity with misspelled, translated and transliterated words as correct.
The final form of the guidelines is shown in Appendix \ref{app:guidelines}.

Ten annotators (1 for the trial, and 9 for the main study) were recruited from our organisation with the condition to be fluent in English.
Each set of 600 queries (DS1, DS2, and DS3) was given to three annotators.
The annotation campaign was performed using Doccano \citep{doccano}.
The guidelines and the annotation tool were presented in a 30-minute workshop where annotators could ask questions. 
They could consult the guidelines and  contact the researchers if they needed any clarification during the experiment too.
After, one week was set aside for each annotator to complete the annotations individually.

\begin{table*}
\begin{small}
    \centering
    \begin{tabular}{c|cccccc||ccc}
        & \textbf{Artist$_{t}$} & \textbf{Artist$_{u}$} & \textbf{WoA$_{t}$} & \textbf{WoA$_{u}$} & \textbf{\%query$_{w/oents.}$} &  \textbf{ents./query} & \textbf{Train} & \textbf{Pre-train} & \textbf{Human} \\ \hline
        \textit{DS1} & 303 & 289 & 208 & 202 & 58\% & 2.0 $\pm$ 1.0 & 15\% & 51\% & 29\% \\
        \textit{DS2} & 285 & 271 & 221 & 220 & 56\% & 1.9 $\pm$ 0.9 & 14\% & 43\% & 30\% \\
        \textit{DS3} & 299 & 284 & 229 & 229 & 57\% &  2.0 $\pm$ 1.1 & 15\% & 44\% & 24\% \\
        \textit{Trial} & 383 & 360 & 270 & 269 & 56\% & 2.0 $\pm$ 1.0 & 11\% & 47\% & 27\% \\
    \end{tabular}
    \caption{Part I shows the total ($\text{[Type]}_{t}$) and unique ($\text{[Type]}_u$) numbers of \textit{Artist} and \textit{WoA} mentions; \% of queries with no entities (\%query$_{w/oents.}$); and per query mean and std. of entity mentions (ents./query). Part II shows \% of unique \textit{test} entities in the \textit{train} set (Train), seen during model \textit{pre-training} (Pre-train) or known to \textit{humans} (Human).}
    \label{tab:ds_stats}
    \end{small}
\end{table*}

\subsection{Ground-truth \textit{MusicRecoNER} Corpus}
\label{sec:groundtruth}
Often in related works, a ground-truth corpus is obtained by using full agreement or majority voting \citep{nangia-bowman-2019-human,lin-etal-2020-triggerner} (e.g. tag named entities on which at least two out of three human annotators agreed).
However, here, because we wanted to establish a human baseline and have a corpus exhaustively annotated, we labelled the ground-truth corpus ourselves from scratch.

Compared to the settings of the human subject study, we had access to the original Reddit post titles including capitalised text and punctuation.
During the annotation, we used web and music streaming search engines to check if certain entities were \textit{Artist} or \textit{WoA}.
The full Reddit post was also used to disambiguate cases when a name could be both an \textit{Artist} or a \textit{WoA}.
The most challenging examples were discussed among us.
The ground-truth preparation together with the adjudication discussions happened over several weeks, as the process to disambiguate entities was more complex.

Statistics about each dataset are presented in Part I of Table \ref{tab:ds_stats}.
\textit{Artist} mentions are more common than \textit{WoA} mentions.
Regardless of the type, we could notice that a large majority of entity mentions are unique in each dataset.
The mean number of entity mentions per query is around $2$, with the maximum varying between 6 and 10.
From these, the proportion of queries with no entity is on average $56\%$.

\section{Evaluation protocol}
\label{sec:experiments}

\subsection{Fine-tuned Transformer Baselines}
\label{sec:baselines}
The goals of the human subject NER study are to establish a human baseline on this challenging dataset of noisy queries for music recommendation and to learn from the human linguistic behavior in comparison to the most common NER systems nowadays, the fine-tuned transformers.
We consider three language models proven to have good results in various natural language tasks including language understanding, sequence labeling or text classification: BERT \citep{devlin-etal-2019-bert}, RoBERTa \citep{liu2019roberta} and MPNet \citep{Song_2020}.

\textbf{BERT} \citep{devlin-etal-2019-bert} is a multi-layer bidirectional encoder based on the original Transformer architecture \citep{Vaswani2017}.
It is pre-trained on: 1) the cloze task, i.e. to predict a masked token from the left and right context; and 2) next sentence prediction, i.e. to predict the next sentence from a given one.
\textbf{RoBERTa} \citep{liu2019roberta} has the same architecture as BERT, but incorporates multiple training steps proven to lead to an increased performance than the original model: the training of the model using more data, with larger batches, on longer sequences and for a longer time; 
and keeping only the cloze task as a pre-training objective while applying a dynamic masking schema to the input training data.
\textbf{MPNet} \citep{Song_2020} proposes a new pre-training objective by integrating the masked language modeling objective of BERT and the permuted language modeling objective introduced in XLNet \citep{Yang2019}.
That is, it models the dependency among the masked tokens at prediction (i.e. takes into account the already predicted masked tokens to generate the current one), while providing visibility on the position information of the full sentence (i.e. the positions of the masked token and the next ones to be predicted). 

We fine-tune the pre-trained versions of these models released in the \textit{huggingface transformers} library \citep{wolf-etal-2020-transformers} for token classification / sequence labeling.
We took the largest available version for each of them: \textit{bert-large-uncased}, \textit{roberta-large}, 
and \textit{mpnet-base}.
From all, only BERT is pre-trained on uncased text.

During experiments, we noticed that the model initialisation had a large impact on the results.
This instability is well-documented in the past work, especially when the corpus for fine-tuning was small \citep{zhang2021revisiting}. 
Thus, to overcome bad initialisation and have more coherent results over different runs, we re-initialized the last layer of each pre-trained model.
This also led to faster convergence and more efficient fine-tuning.
We also tried to increase the number of the re-initialized layers to $2$, but the results were similar or sometimes worse.

\subsection{Evaluation Metrics and Schemes}
\label{sec:eval_metrics}

\begin{table*}
\begin{small}
    \centering
    \begin{tabular}{cc|cc|ccc}
     \multicolumn{2}{c|}{\textbf{Ground-truth}} & \multicolumn{2}{c|}{\textbf{Predicted}} & \textit{Strict} & \textit{Exact} & \textit{Type} \\ \hline 
\textit{Artist} & the beatles & \textit{Artist} & the beatles & $o_{c}$ & $o_{c}$ & $o_{c}$ \\ 
\textit{Artist} & the beatles & \textit{WoA} & the beatles & $o_{i}$ & $o_{c}$ & $o_{i}$ \\ 
\textit{Artist} & the beatles & \textit{Artist} & beatle & $o_{i}$ & $o_{i}$ & $o_{c}$ \\ 
\textit{Artist} & the beatles & \textit{WoA} & beatle & $o_{s}$ & $o_{s}$ & $o_{s}$ \\ 
& & \textit{WoA} & love & $o_{s}$ & $o_{s}$ & $o_{s}$ \\
\textit{Artist} & 
the beatles & & & $o_{m}$ & $o_{m}$ & $o_{m}$ \\ 
    \end{tabular}
    \caption{Example of outcomes under various evaluation schemes.}
    \label{tab:evalschema}
\end{small}
\end{table*}

Precision (P), recall (R) and F1 are commonly used to evaluate automatic NER systems \citep{yadav-bethard-2018-survey}.
In our evaluation, we extend these metrics to support a more detailed benchmark and understanding of the kind of errors a NER system makes.
Namely, we also allow for a relaxed system's evaluation, when either segmentation or typing is correct, but not necessarily both.

A NER system can produce various types of outcomes
\citep{Chinchor1993,chinchor-1991-muc}. 
Inspired by this and \citet{Batista2018}, all NER outcomes, which we denote $O$, can be:

\begin{itemize}
    \item \textit{Correct} outcomes ($O_c$): predicted and ground-truth entities match.
   \item \textit{Missing} outcomes ($O_m$): system entirely fails to spot a ground-truth entity.
    \item \textit{Spurious} outcomes ($O_s$): false entities are produced by the system. 
    \item \textit{Incorrect} outcomes ($O_i$): predicted and ground-truth entities do not match because of either typing or segmentation errors.
\end{itemize}

To classify the predictions of a NER system in these categories, we first need to fix \textit{an evaluation schema}.
The most common one in the literature is the \textit{Strict} match \citep{uzzaman-etal-2013-semeval,chinchor-1991-muc} when both segmentation and typing are correct.
Under the \textit{Strict} schema, a prediction is \textit{incorrect} when its boundaries were correct but not its type, or when its type was correct but not its boundaries.
All other cases (e.g. partial segmentation with incorrect type) are classified as \textit{spurious}.

The \textit{Exact} schema classifies a prediction as \textit{correct} when its boundaries match those of the ground-truth, regardless of its type.
In contrast, the \textit{Entity} schema classifies a prediction as \textit{correct} when its type matches that of the ground-truth, regardless of its boundaries.
For these latter schemes, \textit{incorrect} is adapted from its definition in \textit{Strict};
\textit{missed} and \textit{spurious} are the same too.

We use another class of outcomes, \textit{partial} ($O_p$), only when computing the human performance.
As described in Section \ref{sec:humanner}, humans could annotate a text as \textit{Artist\_or\_WoA\_deduced}.
Thus, whenever a human prediction had this label and matched exactly the boundaries of the ground-truth entity, \textit{partial} was incremented and contributed to the final scores with a factor of $0.5$ \citep{chinchor-sundheim-1993-muc}, as follows:

\begin{align}
     R = (|O_c| + 0.5 * |O_p|) / (|O| - |O_s|) \\
     P = (|O_c| + 0.5 * |O_p|) / (|O| - |O_m|)
\end{align}

\noindent We exemplify the different outcomes under the mentioned schemes in Table \ref{tab:evalschema}.

One practical detail regarding the calculation of the evaluation metrics is that we had to apply some segmentation corrections before, to cover the situations when human annotations started or finished in the middle of a word.
This could appear because Doccano did not force automatically an alignment to a desired tokenization (entire words).
Thus, we corrected the start or end index of the concerned span by moving them to the left or right, based on a simple heuristic with regard to the closest found separating character (space or newline) to the concerned word.
We did not intervene when an entity was composed of multiple words and only a part of them were annotated, but we captured this type of errors with the used evaluation schemes.
No correction was needed in the case of model annotations as, during fine-tuning, we propagated the label of the first word token to the rest;
hence, the labels were always consistent for all word tokens.

\subsection{Evaluation Scenarios}
\label{sec:scenarios}
We explicitly consider the novelty of entities.
In the case of humans, this was encoded in the annotation process as we introduced the labels suffixed with \textit{\_known}. 
Fine-tuned models could have seen music entities from the test set during pre-training, when they were exposed to a large amount of unlabelled data
or during fine-tuning, if the train and test sets had common entities.
While this latter exposure could be easily checked, the pre-training exposure is more challenging to assess as it requires access to the pre-training data or to find other ways to test exposure based on the model only \citep{epure2022probing,tanzer-etal-2022-memorisation}.

The solution we adopted targeted BERT, which performed on par with the other models as revealed in Section \ref{sec:results}.
BERT is pre-trained on Wikipedia and BookCorpus \citep{devlin-etal-2019-bert}.
Thus, music entities could be found more likely in the Wikipedia content.
However, some music entities could be quite rarely mentioned in Wikipedia compared to others.
To quantify BERT's exposure to an entity $e$ we used the following method. 
First, we tried to link each entity to Wikipedia by querying the Wikidata knowledge base \citep{Vrandecic2014}.
We re-ranked the returned results to give priority to music entities and returned the first entity whose type was in a pre-defined type list (see Appendix \ref{app:typelists}). 
Second, we computed exposure by adapting the metric proposed by \citet{Carlini2019Secret}:

\begin{equation}
    \label{eq:exposure}
    \text{expo}(e) = \left\{
	\begin{array}{ll}
	\log{|\mathcal{S}|}-\log{\text{rank}(e)} & \textnormal{$e \in $ Wiki.} \\
	0 & \textnormal{$e \not\in $ Wiki.} \\
	\end{array}
	\right.
\end{equation}

\noindent where $\mathcal{S}$ represents all Wikipedia named entities and the function \textit{rank} considers entity popularity (higher the popularity, lower the rank).
We retrieve $\mathcal{S}$ and entity counts from Wikipedia2Vec \citep{yamada-etal-2020-wikipedia2vec}. 
We manually checked the linking for 300 random entities.
$82\%$ were correct, either linked or not found on Wikipedia correctly.
$14\%$ were linked to music-related entities but not the right ones and the rest were errors or missed entities.
Examples of entities with high exposure values are: \textit{the beatles}, \textit{elvis}, \textit{pink floyd}, \textit{metallica}, \textit{drake}, \textit{johnny cash}, \textit{eminem}, \textit{nirvana}, and \textit{coldplay}.
We could notice that all are of type \textit{Artist}.

\section{Results and Discussion}
\label{sec:results}
\begin{table}
\begin{small}
    \centering
    \begin{tabular}{c|c|c|c}
        \textbf{Model} &  \textbf{Artist} & \textbf{WoA} & \textbf{Macro} \\\hline
       BERT & 0.80 $\pm$ 0.03 & 0.72 $\pm$ 0.04 & 0.76 $\pm$ 0.03 \\
       RoBERTa & 0.77 $\pm$ 0.01 & 0.71 $\pm$ 0.05 & 0.74 $\pm$ 0.03 \\ 
       MPNet & 0.80 $\pm$ 0.03 & 0.72 $\pm$ 0.05 & 0.76 $\pm$ 0.04  \\ 
    \end{tabular}
    \caption{\textit{F1} scores under the \textit{strict} evaluation schema.}
    \label{tab:modelchoice}
    \end{small}
\end{table}

        
    
        

We report scores using 4-fold cross-validation on the datasets presented in Table \ref{tab:ds_stats}.
Means and standard deviations (std.) are computed over different folds, different initialisation seeds for the model, and different human annotators.
In most cases, this was over 12 data points as, for each model, the results were aggregated over each dataset as a test and $3$ different initialisation seeds\footnote{All the models were trained and tested on the ground-truth datasets, and did not consider annotator-specific sets.} and 
for the human evaluation, over each dataset as a test and $3$ human predictors per dataset.

When comparing BERT and the other models in Table \ref{tab:modelchoice}, BERT and human baselines in Tables \ref{tab:tab:results_f1} and \ref{tab:tab:results_pr}, and results on Seen versus Unseen entities obtained either by humans or BERT in Table \ref{tab:seen_vs_rare}, scores in bold are statistically larger (p-value$=0.05$).
We test statistical significance with the Mann-Whitney U Test (Wilcoxon Rank Sum Test, \citealt{Mann1947}), which assesses under the null hypothesis that two randomly selected observations X and Y come from the same distribution.

\subsection{Fine-tuned Transformer Baselines}
\label{sec:comparison_models}
Table \ref{tab:modelchoice} shows that the fine-tuned BERT, pre-trained on uncased text, and MPNet yield the largest \textit{F1} scores for each entity type or overall.
RoBERTa is statistically comparable and only marginally lower than the other models.
Although MPNet and RoBERTa share the same pre-training corpus and the Transformer architecture, the addition of the permuting language objective to the cloze task gives a slight advantage to MPNet.
We use BERT for the rest of the experiments.

\subsection{Humans vs. Fine-tuned BERT}
\begin{table}
    \begin{small}
    \centering
    \begin{tabular}{c|c|c|c}
      & \multicolumn{3}{c}{\textbf{Artist}} \\ 
    & Strict & Exact & Entity \\ \hline
      BERT  & 0.80 $\pm$ 0.02 & 0.84 $\pm$ 0.02 & 0.83 $\pm$ 0.02  \\
      human & 0.77 $\pm$ 0.06 & 0.84 $\pm$ 0.05 & 0.81 $\pm$ 0.05  \\ \hline
      & \multicolumn{3}{c}{\textbf{WoA}} \\
      & Strict & Exact & Entity \\ \hline
      BERT  &  0.71 $\pm$ 0.04 & 0.75 $\pm$ 0.04 & 0.78 $\pm$ 0.04 \\ 
      human & 0.74 $\pm$ 0.07 & 0.79 $\pm$ 0.07 & 0.80 $\pm$ 0.05 \\
    \end{tabular}
    \end{small}
    \caption{\textit{F1} scores under different evaluation schemes.}
    \label{tab:tab:results_f1}
\end{table}

\begin{table}
\centering
    \begin{small}
        \begin{tabular}{c|c|c|c}
       & & P & R\\ \hline
        
       \textbf{Artist} &  BERT  & 0.79 $\pm$ 0.02 & \textbf{0.82 $\pm$ 0.03}   \\
       & human & \textbf{0.82 $\pm$ 0.04} & 0.73 $\pm$ 0.07  \\  \hline
        
       \textbf{WoA} & BERT  & 0.67 $\pm$ 0.04 & 0.74 $\pm$ 0.05 \\
       & human & \textbf{0.78 $\pm$ 0.07} & 0.70 $\pm$ 0.08 \\ 
    \end{tabular}
    \end{small}
    \caption{\textit{P}recision (P) and and \textit{R}ecall (R) under the \textit{strict} evaluation schema.}
    \label{tab:tab:results_pr}
\end{table}
Table \ref{tab:tab:results_f1} shows that the performance of BERT is comparable to that of the human baseline in terms of \textit{F1} score.
However, Table \ref{tab:tab:results_pr} shows that humans and BERT perform differently in terms of \textit{precision} and \textit{recall}.
Humans have a higher \textit{precision}, for both \textit{Artist} and \textit{WoA}, whilst BERT has a marginal or significantly larger \textit{recall} than humans, especially for \textit{Artist}.
We confirmed that this phenomenon was not due to a particular precision / recall compromise by testing various precision / recall value and optimizing on F1.
Also, BERT has a lower \textit{precision} than the \textit{recall}, but we see the opposite for humans.
Considering Equations 1 and 2, the model appears to hypothesize spurious entities more often, while humans tend to miss entities more often.

Table \ref{tab:tab:results_f1} also shows that the \textit{F1} scores under \textit{Exact} and \textit{Entity} schemes are larger than under \textit{Strict} as some of the errors produced are because of segmentation or typing.
However, we can notice a different behaviour for the two entity types for both BERT and humans.
In the case of \textit{WoA}, the \textit{Entity} \textit{F1} scores are slightly larger than those obtained under the \textit{Exact} schema, showing that boundary errors happen more frequently.
On the contrary, for \textit{Artist} entities, the segmentation is more often correct, but the typing is wrong.

\subsection{Error Analysis}

\begin{figure*}
    \centering
\includegraphics[width=1\textwidth]{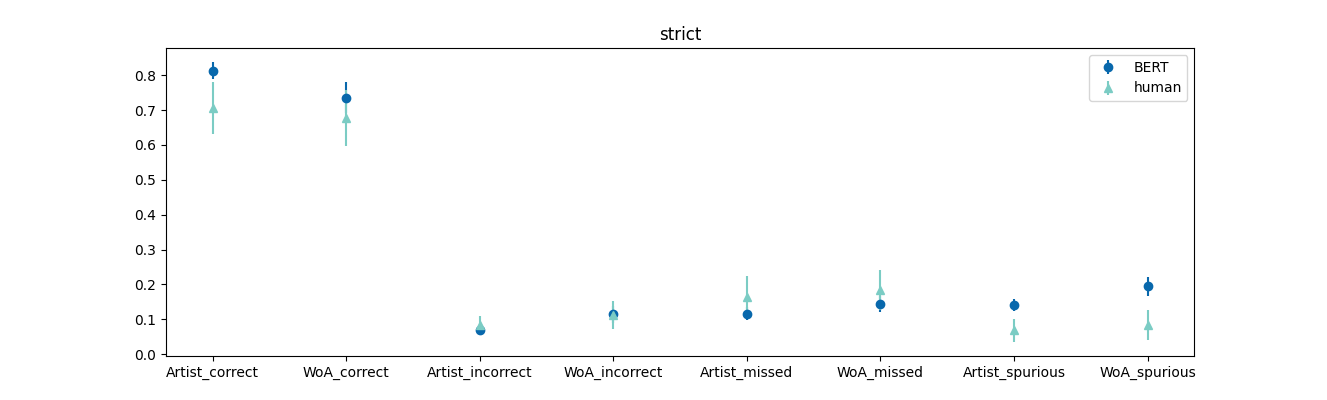}
    \caption{Normalized \textit{correct}, \textit{incorrect}, \textit{missed} and \textit{spurious} outcomes per entity type under \textit{strict}.}
    \label{fig:error_analysis}
\end{figure*}
Figure \ref{fig:error_analysis}, showing a detailed error analysis, confirms that indeed BERT has more often \textit{spurious} outcomes than humans, for both entity types.
Also, humans miss to annotate ground-truth entities more often than BERT.
We can equally observe that BERT is highly superior in identifying correct named entities.
Previous works on NER \citep{lin-etal-2020-triggerner,epure2022probing} have discussed that a system should learn to exploit the context \ (i.e. the non-entity words) rather than entity memorisation to generalise.
However, the high number of correctly recognised entities as well as the frequent spurious entities suggest that this may not be the case here; 
and BERT's behaviour may be linked to entity exposure. 
As shown at the end of Section \ref{sec:scenarios}, the entities with the highest exposure score were of type \textit{Artist}.
We could see in Figure \ref{fig:error_analysis} that there are a lot more correct \textit{Artist} entities, and the number of \textit{missed} and \textit{spurious} outcomes for \textit{Artist} is lower, which seems to be aligned with our hypothesis related to entity exposure.

\subsection{Impact of Entity Exposure}
In Table \ref{tab:ds_stats}, Part II, we show the percentage of entities known by at least one annotator among the three in each dataset.
This varies between 24\% and 30\%.
In practice, each annotator has known at most this number of entities, which confirms that most entities from the collected corpus were new to our subjects.
The entity exposure is much larger for BERT.
While the train and test sets share only maximum $15\%$ of the entities, BERT has seen up to a half of corpus' entities during pre-training.

To check the model's performance on seen versus unseen entities, we show \textit{Recall} scores for these groups in Table \ref{tab:seen_vs_rare}.
\textit{Seen} entities are those present in the train set or with $expo(e)>1$. 
\textit{Unseen} entities have $expo(e)=0$ and are not known to humans.
The rest of the entities are discarded from the evaluation.
BERT's recall on \textit{Seen} is much larger than on \textit{Unseen}, which confirms our hypothesis that memorisation plays a role.
However, the model seems to rely significantly on context too given that the results on Unseen are still quite high.

\begin{table*}
\begin{small}
    \centering
        \begin{tabular}{c|c|c||c|c}
      & \multicolumn{2}{c}{\textbf{Artist}} & \multicolumn{2}{c}{\textbf{WoA}} \\
        
     & Seen & Unseen  & Seen & Unseen \\ \hline
      BERT  & \textbf{0.86 $\pm$ 0.03} & 0.74 $\pm$ 0.05 & \textbf{0.81 $\pm$ 0.05} & 0.69 $\pm$ 0.06  \\
      human & \textbf{0.77 $\pm$ 0.07} & 0.63 $\pm$ 0.08 &  \textbf{0.74 $\pm$ 0.08} & 0.66 $\pm$ 0.09 \\
    
    \end{tabular}
    \caption{\textit{R}ecall scores under the \textit{strict} evaluation schema on Seen and Unseen.}
    \label{tab:seen_vs_rare}
    \end{small}
\end{table*}

We also report the results of humans in Table \ref{tab:seen_vs_rare} and see a similar pattern.
Although the split is made considering \textit{the model's exposure}, humans are also very likely to know entities from \textit{Seen}.
The lower humans' scores on \textit{Unseen} show that the recognition of these entities is quite challenging, possibly because of insufficient context.
For example, "songs bands similar to sales getting it on off and on and porches mood" contains an enumeration that is difficult to segment and type (\textit{Artist}: "sales", "porches"; \textit{WoA}: "getting it on", "off and on", "mood").
Also, entity typing is ambiguous in "anything similar to some people say" (\textit{WoA}).
For these imperfectly recognised entities, including external resources such as gazetteers or search engines might be an option to explore.

\section{Conclusion}
\label{sec:conclusion}
In this work, we investigated the human linguistic behavior when performing NER in the music domain.
We created \textit{MusicRecoNER}, a new corpus of complex noisy queries for recommendations annotated with \textit{Artist} and \textit{WoA} entities.
We then designed and conducted a human subject research study to establish a human baseline and learn from its comparison with the most popular systems nowadays, fine-tuned transformers.
We performed a thorough evaluation covering multiple metrics, schemes and scenarios, including a careful analysis of the impact of entity exposure on results.

The results obtained by the algorithmic baselines were comparable to the human ones. 
Yet, the detailed evaluation showed that humans yielded a better precision while the model had a better recall, linked also to entity exposure during pre-training and fine-tuning.
Thus, when evaluating fine-tuned pre-trained models, checking their performance on new entities shows their real generalisation ability.
Regarding the NER evaluation protocol, human performances were much better under a more relaxed schema focused on segmentation or typing only.
Such a schema could prove a more realistic setup to aim to when training models too. 
Also, we noticed that the relevant schema depended on the entity type as \textit{Artist} was better segmented, while \textit{WoA} better typed. 

Contrary to previous claims, we show that, in our domain, NER in challenging conditions such as noisy text, and irregular or novel entities is rather hard for humans even when provided with complex instructions and multiple examples.
Thus, although we could learn from the human linguistic behaviour, we should not, by default, assume their results to be a target for any NLP problem. 
For some tasks, it is common when establishing a human baseline to consider it as an upper bound for the model.
This is not necessarily a desirable outcome in our case as it would imply mislabelling $1/3$ \textit{WoA} entities.
More generally, as we also showed by studying the impact of entity exposure, algorithms can store a lot more knowledge than humans and one may want to leverage this as much as possible.

As for proposing a better system to perform music NER, one next step would be to continue the model's pre-training on more related data, in our case music, to get even more exposure, or to integrate gazetteers.
Still, given the rate of new entities in our domain, forcing the model to rely more on context, when context is not confusing, is another desirable future direction.
In case of context ambiguity, asking questions to clarify the request and supporting user interaction in  natural language could be ultimately the answer towards a more suitable, but still very challenging solution.
We plan to explore these ideas as future work.

\section{Limitations}
\label{sec:limitations}
We further discuss the limitations of our work.
The corpus of noisy complex queries in natural language we use in the human subject study and we release is built based on a single source, Reddit.
The demographics of the users using Reddit are relatively narrow, with a majority being male, young, and educated\footnote{\url{https://foundationinc.co/lab/reddit-statistics/}}. 
Moreover, users seeking music recommendations on this type of forums may be rather "music enthusiasts" and may not represent regular music listeners.
The implications are that the language employed in these queries could be specific to this category of population.
Also, the mentioned entities could reflect the music taste of this type of profiles only.  
This latter implication turned to be an advantage for us as we ended up with many novel entities, unknown by the annotators who participated in the study.
As for the first implication, we manually checked the queries, and found them quite diverse, not necessarily using a specific vocabulary but more general language expressions. 
An alternative to creating such a corpus could have been a Wizard of Oz experimental setup \citep{Green1985}.
However, this would require significantly more costs and would highly depend on the type of profiles interested in participating in such a music discovery experiment.

Second, we pre-processed the corpus in order to simulate written or transcribed speech-based human-computer interactions.
However, the steps we took may be largely insufficient to simulate the kind of noise found in transcriptions.
As we also discussed in Section \ref{sec:datapreprocessing}, we did not inject any artificial noise for named entities, while spelling errors when automatically transcribing them are a common problem. 
Another limitation regarding named entities is the computation of the model's exposure by leveraging Wikipedia.
Our linking was quite rudimentary and imperfect, as we reported in Section \ref{sec:scenarios}.
Moreover, for retrieving entity ranks, we used Wikipedia2Vec \citep{yamada-etal-2020-wikipedia2vec}, which is built on a slightly older Wikipedia version than the one BERT was trained on.
Therefore, the results obtained by the model on the \textit{Unseen} dataset may be slightly larger, as the model could have been exposed to some of these entities.
However, our goal was to prove a phenomenon---the impact of named entity exposure on the results, even if this impact may be marginally underestimated.

Finally, the annotators recruited from our organisation have similar age and demographics.
Also they likely have a richer musical background compared to regular human subjects.
This signifies that, in reality, the number of novel entities could be higher, which could also impact the overall results obtained with the human baseline.
Nevertheless, this hypothesis could be tested only by running subsequent studies including more subjects.

\section{Ethical Considerations}
\label{sec:ethics}
We have provided most of the details about data collection, data cleaning and pre-processing, and the annotation procedure and guidelines in Section \ref{sec:corpus} and Appendices.
We discuss further various ethics-related aspects not covered yet in the paper.

The dataset was gathered from the music suggestion subreddit via the Reddit API.
According to the privacy policy of Reddit\footnote{\url{https://www.reddit.com/policies/privacy-policy}}, third parties can freely access public content via the API.
We have not gathered any other information besides the public posts---their titles and descriptions.

As previously mentioned, the annotators were recruited from our organisation.
They performed the annotation tasks during their regular paid hours.
Moreover, the participation was fully on voluntarily basis, following a public call for participation by the authors within the organisation.



\bibliography{anthology,custom}
\bibliographystyle{acl_natbib}

\appendix

\clearpage

\section{Data Filtering Keywords}
\label{app:filterkeywords}
The list of keywords used in the data cleaning and pre-processing steps are presented in Table \ref{tab:keywords}. 
These keywords were used to filter out posts, which were manually verified after. The outcome of the verification was either to exclude these posts from the data, or to keep the posts as they were or after having removed specific words (as described in Section \ref{sec:datapreprocessing}). 
We have considered both lower and upper case variations of each keyword. 

\begin{table}
\begin{small}
    \centering
    \begin{tabular}{c|c|c}
       Spotify & Playlist & Radio \\
       Check out & Skunk & YouTube \\
       http & Buy & Enjoy \\
       Compiled & Zoom & Download \\
       Event & Promotion & Quarantine \\
       Weekly & Someone & Anybody \\ Instagram & iTunes & Playlist \\
       $[$ & \{ & /r/ \\
       Hour & Release & Stream \\
       Official & Video & Anyone \\ 
       Guys & S**t & F**k \\
       Reddit & Apple & Link \\
       Post & Soundcloud & Radio \\
       Made & Our & Thanks \\
       Hi & Hello & Mix \\
       Listen & Cover & My \\
       If & You & Inside \\
       Tell me & Check & Example \\
       Text & &
    \end{tabular}
    \caption{Keywords used for manually filtering out Reddit posts in the data pre-processing step.}
    \label{tab:keywords}
    \end{small}
\end{table}

\section{Annotation Guidelines}
\label{app:guidelines}

\subsection{Introduction}
The goal of this annotation experiment is to identify names of artists (e.g. bands, singers, composers) and names of works of art (e.g. albums, tracks, playlists, soundtracks, movies, video games) in music-related requests. 
The requests could be single- or multi-line. Also, they are unformatted, meaning that they contain no capitalized letters or punctuation marks. Also contractions such as "\textbf{Artist’s} first album", "\textbf{don’t}" are written as if pronounced, specifically "artists first album" and "dont".

Through this experiment, we study how well humans can identify named entities (artists and works of art) in unformatted text by relying on the \textit{request content only}, and on one's own knowledge.
For this reason, it is important that during annotation you do not consult the Internet to verify if some parts of text are named entities, but rely on your intuition after reading the text. 

\subsection{Named Entity Categories}
There are two categories referring to the entity type \textit{Artist}; two categories referring to the entity type Work of Art (\textit{WoA}); and one category for dealing with ambiguous cases as follows:

\paragraph{1. Artist\_known.} 
This category should be used for sequences of words denoting an artist that is previously known to the annotator.

In the next request I recognize "queen" and "the clash" to be \textit{Artist} entities because I knew them from the past:
\textit{i like \textbf{queen} and \textbf{the clash} what else should i listen to}.

\noindent Note that when "the" is part of the name (e.g. "the clash"), it should be annotated likewise.

\paragraph{2. Artist\_deduced} This category should be used for sequences of words denoting an artist that is not known to the annotator, but deduced from the text. 

In the next request I recognize "stephan forté" to be an \textit{Artist}: \textit{looking for something like the first album of \textbf{stephan forté}}.

\noindent I have never heard of this artist before, but I deduced it from the request’s content.

\paragraph{3. WoA\_known}
This category should be used for sequences of words denoting a work of art that is previously known to the annotator.

In the next request I recognize "karma police" to be a \textit{WoA} because I knew it before: \textit{im a very picky music listener but i love \textbf{karma police} any other suggestions}.

\paragraph{4. WoA\_deduced}
This category should be used for sequences of words denoting a work of art that is not known to the annotator, but deduced from text.

In the next request I recognize "special affair" to be a \textit{WoA}: \textit{songs like \textbf{special affair}}.

\noindent I have never heard of this work of art before, but I deduced it from the request’s content.
	
\paragraph{5. Artist\_or\_WoA\_deduced}
This category is used for sequences of words recognised to denote an artist or a work of art, but choosing between the two entity types is challenging.
	
In the next request I recognize "superunloader" to be either an \textit{Artist} or a \textit{WoA}:\textit{ music like \textbf{superunloader}} 

\noindent I have never heard of this before and it is difficult for me to distinguish between the two options.

\subsection{(Challenging) Examples}
Please read the following examples carefully and re-consult them during the experiment whenever needed.

\paragraph{Relevant named entities not related to music.}
A text could contain other types of works of art such as movies or video games. Annotate these names using the category \textit{WoA}. 
Similarly, annotate with the \textit{Artist} category movie directors, filmmakers, music composers and so on. 
All the other types of named entities not related to \textit{Artist} and \textit{WoA} must be ignored (e.g. names of countries, music genres etc.).

In the example below, "gemini" is annotated as \textit{WoA} and "ang lee" as \textit{Artist}:
\textit{i recently watched this film \textbf{gemini} made by \textbf{ang lee} and liked the soundtrack any similar recommendations of this}.

\paragraph{Multiple named entities clearly delimited.}
A text could contain multiple entities which are clearly delimited by other words such as "by", "from", "and" etc. Please annotate all of them individually.

In the example below, "heartbeat" is a \textit{WoA} and "annie" is an \textit{Artist}: \textit{songs with similar vibe and structure as \textbf{heartbeat} by \textbf{annie}}.

In the example below, "hallelujah" is a \textit{WoA} and "jeff buckley" is \textit{Artist}:
\textit{other beautiful songs by \textbf{jeff buckley} apart from \textbf{hallelujah}}.

\paragraph{Multiple named entities with no delimitation.} A text could contain multiple entities which are not clearly delimited. Try to annotate each segment of text individually with its corresponding named entity category.

In the example below, if the annotator recognizes the entities, then 3 separate \textit{Artist} entities should be selected, namely "imagine dragons", "bastille", and "daya": \textit{singers bands like \textbf{imagine dragons} \textbf{bastille} and \textbf{daya}}.

However, if not all entities are known from the past, then the unknown span of text could be annotated either as \textit{Artist\_or\_WoA\_deduced}, \textit{Artist\_deduced} or \textit{WoA\_deduced} depending on the content and the annotator's intuition. 
For instance, if the annotator recognizes only "imagine dragons" but not the rest, then "bastille and daya" could be considered either 1 entity ("bastille and daya") or 2 entities ("bastille", "daya") and further annotated with any of the 3 categories mentioned above.

\paragraph{Named entities collated with 's from the possessive case.} 
In this case, include the "s" as part of the named entity. 

In the example below, "toni braxton" is the real name of the artist, but "toni braxtons" (coming from "toni braxton's") is actually annotated as an \textit{Artist} entity:
\textit{newer 2005+ ballads in the style of \textbf{toni braxtons} un break my heart and \textbf{stevie wonders} all in love is fair}.
Similarly for "stevie wonders" (coming from "stevie wonder's"). 

\paragraph{Nested named entities.} 
A text could contain nested entities. 
This means that there is a larger text segment that could be considered as an entity and a smaller text segment inside the larger one that could be also considered as an entity.

In this case, always favor the \textit{innermost} text segment with an exception which is described below. Multiple examples are given further.

In the example below, "treasure planet" is annotated as \textit{WoA} and not "treasure planet soundtrack" (which is also a \textit{WoA}, but the innermost one is chosen):
\textit{looking for calm violin music very similar to the first 34 seconds of 12 years later from the \textbf{treasure planet} soundtrack}.

In the example below, "ezra collective" and "ty" are annotated as 2 separate \textit{Artist} entities and not as one: "ezra collective feat ty":
\textit{recommend me some good jazz hip hop songs with rap like chapter 7 by \textbf{ezra collective} feat \textbf{ty}}.
\noindent There is also a third entity, "chapter 7", annotated as \textit{WoA}):

In the example below, although "i took a pill in ibiza seeb remix" could be considered as a \textit{WoA}, the innermost entities are annotated instead, namely "i took a pill in ibiza" as \textit{WoA} and "seeb" as \textit{Artist}: \textit{songs similar to \textbf{i took a pill in ibiza} \textbf{seeb} remix}.

Exception: if the name of a well-known band that you recognize is composed of 2 or more individual artist names, then annotate the band name using the category \textit{Artist\_known}. In the example below, I recognized that "emerson lake and palmer" is the name of a band despite the fact that it refers to three individual artists ("emerson", "lake", "palmer"): 
\textit{other artists similar to \textbf{emerson lake and palmer}}.

\paragraph{Explicit versus implicit named entities.} 
There are cases when an \textit{Artist} or a \textit{WoA} are mentioned in text, but these entities are not explicitly named. For instance, neither "the last album", nor "this singer" are explicit named entities in the request below; hence they must not be annotated:
\textit{show me something similar to the last album of this singer}.

\paragraph{(Incorrect) Variations of the original named entities.} 
The text may contain variations of the original names of the entities (including misspelled, missing, translated or transliterated words). 
Normally, in order to recognize an incorrectly written named entity, the named entity must be already known to the annotator. 
In these cases, even if the named entity does not match exactly the real name, the annotator is required to annotate the corresponding span of text using the named entity categories ending with the "\_known" suffix.
 
In the example below, the annotator recognizes "hey ponchuto" to be mistakenly written:
\textit{fast dancey blues or songs like \textbf{hey ponchuto} from \textbf{the mask}}.
The original named entity which the annotator knows from the past is "hey pachuco". 
Thus, "hey ponchuto" is annotated as \textit{WoA\_known}. 
Note that "the mask" is a WoA too (a movie).

\section{Pre-defined Entity Types for Wikidata Linking}
\label{app:typelists}
In order to ensure that the entity linking gives priority to music-related entities, we re-rank the returned results.
Specifically, we return the first entity whose type matches any of the criteria below:
\begin{itemize}
    \item \textit{Artist}:  type matches exactly one of the following types---\textit{musical group}, \textit{rock group}, \textit{supergroup}, \textit{musical ensemble}, \textit{girl group}, or it contains one of the following strings---\textit{band}, \textit{duo}, \textit{musician}, \textit{singer}.
    
    \textit{WoA}:  type matches exactly one of the following types---\textit{album}, \textit{musical work/composition}, \textit{song}, \textit{single}, \textit{extended play}, or it contains one of the following strings---\textit{album}, \textit{song}.
\end{itemize}

If the previous matching fails, the fallback is the first entity of type \textit{human} for an \textit{Artist} entity, or of type \textit{video} or \textit{film} for a \textit{WoA} entity.
If none of these type criteria is met, then an empty string, corresponding to failed linking is returned.

\section{Computational Information}
\label{app:compinfo}
For training and evaluation, we had a $32$-core Intel Xeon Gold 6134 CPU @ 3.20GHz CPU with 128GB RAM, equipped with $4$ GTX 1080 GPUs, each with 11GB RAM.
Fine-tuning a single model on three datasets from the four we annotated during 3 epochs and testing it on the hold-out dataset on a single GPU took about 2 minutes.

\end{document}